\documentclass{sig-alternate}

\usepackage{graphicx}
\usepackage{amsmath}
\usepackage[skip=5pt]{caption}
\usepackage{hyperref}
\usepackage{enumitem}

\newcommand{\pseudosection}[1]{\vspace{0.5\baselineskip} \noindent {\bf #1}}

\begin{document}
% --- Author Metadata here ---
\conferenceinfo{Foundations of Digital Games}{2015 Asilomar Conference Grounds, California USA}
%\CopyrightYear{2007} % Allows default copyright year (20XX) to be over-ridden - IF NEED BE.
%\crdata{0-12345-67-8/90/01}  % Allows default copyright data (0-89791-88-6/97/05) to be over-ridden - IF NEED BE.
% --- End of Author Metadata ---

\title{Monte-Carlo Tree Search for Simulation-based Strategy Analysis}

\numberofauthors{1}
\author{
\alignauthor
Alexander Zook, Brent Harrison and Mark O. Riedl\\
\affaddr{School of Interactive Computing, College of Computing}\\
\affaddr{Georgia Institute of Technology}\\
\affaddr{Atlanta, Georgia, USA}\\
\email{\{a.zook, bharrison6, riedl\}@gatech.edu}
}

\toappear{}

\maketitle
\begin{abstract}
Games are often designed to shape player behavior in a desired way; however, it can be unclear how design decisions affect the space of behaviors in a game.
Designers usually explore this space through human playtesting, which can be time-consuming and of limited effectiveness in exhausting the space of possible behaviors.
In this paper, we propose the use of automated planning agents to simulate humans of varying skill levels to generate game playthroughs.
Metrics can then be gathered from these playthroughs to evaluate the current game design and identify its potential flaws.
We demonstrate this technique in two games: the popular word game {\it Scrabble} and a collectible card game of our own design named {\it Cardonomicon}.
Using these case studies, we show how using simulated agents to model humans of varying skill levels allows us to extract metrics to describe game balance (in the case of {\it Scrabble}) and highlight potential design flaws (in the case of {\it Cardonomicon}).
\end{abstract}

\category{Applied Computing}{Computers in other domains}{Personal computers and PC applications}[Computer games]

\terms{Measurement}


%%%%%%%%%%%%%%%%%%%%%%%%%%%%%%%%%%%%%%%%%%%%%%%%%%%%%%%%%%%%%

\section{Introduction}

\noindent Creating a game, from a small-scale indie game to a large AAA title, requires careful consideration of how design decisions shape player behavior.
%Designers must make decisions about what the mechanics of the game should be, what challenges should be presented to the player, and how the player will win, just to name a few.
Any single design choice---e.g., including a word in {\it Scrabble}---has rippling consequences for the space of strategies available to players.
While designers directly shape the space of actions available to players, they typically aim to create a core gameplay loop of player behavior \cite{salen2003:rulesplay} and/or balance the competitive elements of a game \cite{elias2012:characteristics-games, jaffe2012:balance}.
Analyzing player strategies is especially important when games afford many levels of play: particularly when highly skilled players may pursue entirely different strategies to amateurs \cite{elias2012:characteristics-games}.
%Each of these decisions affect how players will interact with the game world or the strategies that they will develop to complete the game. 
%As a designer, one wants to have a sense of how players will be playing your game so that one can ensure that players have the best experience possible. 
At the moment, however, designers lack ways to quickly analyze the space of afforded behaviors and how this space changes for players of different skill levels. 

Existing approaches to game analysis rely on playtesting with humans \cite{seifel-nasr2013:game-analytics-book}.
While effective for informing some types of game design, these methods can be expensive and time-consuming.
Further, playtesting is not exhaustive and fails to address many design questions related to the space of {\it possible} play within a game.
As a result, existing methods focus on abstracting limited examples of pre-existing behavior rather than extracting knowledge about behavior within a game domain itself.

Analyzing player strategy requires understanding patterns of player actions in a game \cite{osborn2014:playtrace-metric, osborn2014:trace-eval}.
In competitive games, it is also important to consider how player strategies differ between high- and low-skill players to understand the range of strategic play a design affords.
In this paper, we show how planning technologies can simulate player behavior at a variety of skill levels and how behavior metrics extracted from simulated playthroughs can be used to analyze a game design at multiple levels of abstraction.
We apply Monte-Carlo Tree Search (MCTS) to turn-based adversarial games and develop a taxonomy of metrics to help understand the space of strategic options available to players of varying skill over the course of the game.

While designers are concerned with many aspects of player strategy, we study four categories of player actions: summary statistics, atoms, chains, and action spaces.
{\it Summaries} are high-level metrics of overall gameplay characteristics.
{\it Atoms} are metrics of individual, context-free player actions.
{\it Chains} are metrics about the relationships among sequences of player and inter-player actions \cite{bjork2005:patterns, cadwell2013:counterplay}.
{\it Action spaces} address the range of possible actions over the course of a game \cite{elias2012:characteristics-games}.
%
%we specifically address quantifying the concepts ``game complexity tree'' \cite{elias2012:characteristics-games} and ``counterplay \cite{bjork2005:patterns, cadwell2013:counterplay}.''
%The \textit{game complexity tree} refers to number of actions available to players; both total actions available and total viable actions are important to design \cite{elias2012:characteristics-games}.
%We operationalize game complexity trees in terms of the total number of actions available to a player on a given turn compared to the total number of actions \textit{taken} by an agent with a given level of skill.
%\textit{Counterplay} refers to the possibility for a player to take actions to foil the goals of an opposing player \cite{bjork2005:patterns}.
We operationalize these concepts in terms of the actions available to and taken by simulated agents in a game.
Smith, Nelson, and Mateas \cite{smith2010:ludocore} analyze the effects of simulated actions in a game by generating playtraces when players are constrained to always or never take particular actions under certain conditions.
Jaffe et al. \cite{jaffe2012:balance} develop a framework for ``restricted play'' and evaluate win rate in adversarial games when agents are subject to a variety of action restrictions.
Using playtraces generated by agents with differing computational constraints we approximate human players of differing skills.
We build on these approaches by considering restrictions on agent search capabilities, extending to non-deterministic domains, and providing additional levels of abstraction between concrete exemplar traces \cite{smith2010:ludocore} and abstract summary statistics \cite{jaffe2012:balance}.
We apply our method to two games: the classic word game {\it Scrabble} and a card game we made that models parts of {\it Magic: The Gathering} and {\it Hearthstone}, called {\it Cardonomicon}.

Our work makes three contributions:
\begin{itemize}[noitemsep]
\item Demonstrating how Monte-Carlo Tree Search (a planning algorithm) can simulate player behavior with varying player capabilities in a game.
\item Providing four levels of design metrics to organize play behavior analysis.
\item Case studies applying this approach to two sample games: {\it Scrabble} and {\it Cardonomicon}, a card game based on common trading card game mechanics.
\end{itemize}
Through this work we aim to expand the use of simulated agents and related techniques to inform the analysis and design of games.
These methods can enhance human design practices, support a hypothesis-driven scientific study of game design, and enable the automated generation of games.

%%%%%%%%%%%%%%%%%%%%%%%%%%%%%%%%%%%%%%%%%%%%%%%%%%%%%%%%%%%%%

\section{Related Work}
Automated game analysis is a growing field concerned with developing methods to understand a game design without requiring human play \cite{jaffe2012:balance, nelson2011:analytics-noplayers, smith2010:ludocore, togelius2011:sbpcg}.
Researchers have developed methods to simulate game systems with little or no human player activity \cite{dormans2009:machinations}, use constraint solving to check for the existence of traces consistent with speculative assumptions \cite{smith2011:asp-pcg}, apply exhaustive search to constrained game spaces \cite{sturtevant2013:bfs-design}, and use game theory to evaluate balance in adversarial games \cite{jaffe2012:balance}.
These efforts typically only evaluate game design properties based on reachability or balance, offering a limited view on how design decisions affect the overall play experience.
Our work provides additional levels of abstraction to understand features of strategies including action distributions and common action patterns.

Automated game generation researchers have simulated player behavior using search to reach goal states \cite{cook2012:coopcoevo, smith2011:tanagra, zook2014:mech-gen-aaai} (possibly subject to constraints on the states visited in the reachability check \cite{smith2013:quantify-play}) and hand-coded heuristics for learnability \cite{togelius2008:gamegen} or design aesthetics such as balance \cite{browne2010:ludi, nelson2015:pcg-book-rules-mechanics}.
By contrast, games user research and game analytics \cite{seifel-nasr2013:game-analytics-book} emphasizes aggregate properties of sets of player behaviors in a game, such as qualitative patterns in player solution strategies \cite{linehan2014:puzzle-pace}, metrics on game length or diversity of options \cite{elias2012:characteristics-games}, or clusters of types of players \cite{sifa2013:tombraider-evol, thurau2009:cnmf-wild}.
Ideally, automated design analysis tools should facilitate reasoning on aggregate properties of player behavior in a game without requiring exhaustive user testing.
As many game state spaces cannot be exhaustively sampled, there is a need for methods to sample potential player behaviors in a game \cite{smith2013:playtrace}.

Evaluating potential player behavior for an arbitrary game design is a central concern of general game playing research \cite{genesereth2005:general-game-playing, love2008:ggp-spec}.
Recently, Monte Carlo Tree Search (MCTS) has emerged as a popular technique in general game playing after the successful application of MCTS to the game of {\it Go}~\cite{browne2012:mcts-survey}.
Game applications of MCTS include: card selection and play in {\it Magic:\ The Gathering} \cite{cowling2012:mcts-magic-imperfect, ward2009:mcts-magic-card}; platformer level completion \cite{jacobsen2014:mcts-mario, tremblay2014:icanjump}; simulations for fitness function heuristics in strategy \cite{mahlmann2011:strat-game-eval}, card \cite{font2013:card-game-gen}, abstract real-time planning \cite{perez2013:map-gen-physical-travelling-salesman}, and general arcade games \cite{nielsen2015:gvg-eval}; and high-level play in board games including {\it Reversi} and {\it Hex} \cite{benbassat2013:mcts-evol}.
MCTS offers the advantages of being game-agnostic, having tunable computational cost, and guaranteeing (eventual) complete exploration of the search space.
Unlike previous uses for (near-) optimal agent play, we use MCTS to sample playtraces in a game while varying agent computational bounds as a proxy for player skill.
Our approach trades exhaustively exploring a small or abstracted design space for sampling larger, non-deterministic game domains, complementing prior work in this area.

%\todo{cut? condense? may have space...}
Game visual analytics researchers have studied ways to visualize game metrics to understand play behaviors \cite{wallner2013:gameplay-viz-analysis}.
Visual analysis examines many game properties, including spatial distributions of events (e.g., using heatmaps), aggregation and identification of player types (e.g., using dimensionality reduction or clustering methods), and summarization of individual player behavior (e.g., through dashboards and in- or out-of game representations).
Here, we focus on representations of the {\it progression of states} in a game---visualization of how gameplay occurs through studying common states, actions, and progressions between states.
A progression of game states can be represented as a sequence of player actions and as a sequence of game states.
State analysis methods have aggregated regularities in states visited in single-player games \cite{andersen2010:state-projection, liu2011:playtracer, wallner2013:gameplay-viz-analysis}, while action analysis methods have studied playtraces as sequences of actions~\cite{osborn2014:playtrace-metric}.
We contribute to action analysis through considering common levels of analysis of the use of actions in games.
Unlike prior work focused on recognizing the similarity of action sequences, we emphasize multiple levels of analysis of how actions are used in a game to inform design decisions.

\section{Methodology}
%The main goal of this paper is to show how game traces generated by simulated agents can be used to extract metrics to evaluate the overall design of a game. 
%In particular, we are interested in examining how players of varying skill levels interact with a game. 

%Understanding the space of possible player strategy in a game requires an example player behavior from a game.

One goal of this work is to automate player strategy analysis.
To do this, we use a simulated agent to sample player behaviors and then describe playtraces using summary statistics and pattern analysis to inform design analysis.
In this section we review our agent simulation technique, Monte-Carlo Tree Search, and describe the game domains we use in case studies to illustrate our approach.
Our studies concern discrete, turn-based, adversarial games that emphasize strategic decision-making.
These are games where two opposing players alternate turns using actions that modify discrete domain values.
Searching over all possible traces in most adversarial game domains is intractable (e.g., evaluating at least all permutations of card draw and play orders between two players), requiring a sampling approach to search.\footnote{Approaches based in abstracting game state or actions are also possible, though these introduce more coarseness in the states or actions checked.}

\subsection{Monte-Carlo Tree Search}
Monte-Carlo Tree Search (MCTS) is a general game playing technique with recent success in discrete, turn-based, and non-deterministic game domains \cite{browne2012:mcts-survey}.
MCTS is a sampling-based anytime planning method that can use additional computational resources to more fully explore a space of possibilities, allowing control over the balance between computational time and exploration of the full space of play behavior.
We choose MCTS as a behavior sampling algorithm for its proven high-level performance, domain generality, and variable computational bounds.
Note that other algorithms may serve this need; we leave comparison to alternative algorithms to future work.
For simplicity, both of our case study domains have been made perfect information to facilitate the use of MCTS.

\begin{figure}[tb]
\centering
\includegraphics[width=\columnwidth]{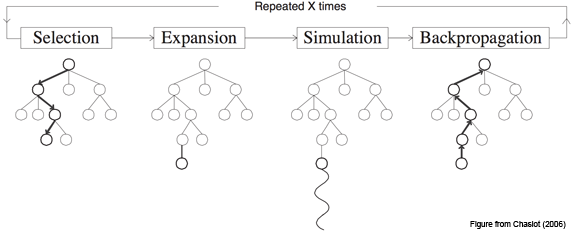}
\caption{Diagram of MCTS algorithm steps.}
\label{fig:mcts}
\vspace{-1.0\baselineskip}
\end{figure}

MCTS's game playing success derives from modeling the quality of a space of actions over the course of a game.
MCTS models game play using a tree to track the value of potential courses of action in a game.
Actions to take are tree nodes and links between nodes indicate the next action(s) available after a prior action (Figure \ref{fig:mcts}).
Nodes for already attempted actions are {\it expanded} and not-yet-attempted nodes are {\it unexpanded}.
Each leaf node in the tree tracks a reward value for the agent depending on if it won or lost the game.
Typically, a reward value of 1 is assigned to wins and a reward value of -1 to losses.

The MCTS algorithm has four steps (Figure \ref{fig:mcts}):
\begin{enumerate}
\item {\bf Selection} Choose how to descend a tree of expanded nodes until reaching an unexpanded node. 
\item {\bf Expansion} Expand the set of actions available at an unexpanded node and choose a new node.
\item {\bf Simulation} Follow a fixed strategy (usually random choice) for how to act over all remaining unexpanded decisions until reaching the end of the game.
\item {\bf Backpropagation} Use the reward from the end game state reached (e.g., win or loss) to update the expected value of the newly expanded node and all of its parent nodes in the tree.
\end{enumerate}
MCTS balances between agents exploring alternative actions and exploiting known good actions.
Typically, selection uses the UCB1 algorithm, which picks a node using a combination of the average reward (eventually) received when taking the action and the proportion of all selections that used that node~\cite{chaslot2006:mcts-go}.
Note that UCB1 forces selection to first visit every possible move at least once before choosing among all visited nodes based on their value.
We use UCB1 because this property ensures our agents fully explore the space of move options before devoting additional resources to better modeling the value of known choices.

In our case studies we use the strength of MCTS to play well in discrete, turn-based, adversarial games and combine this with the ability to tune MCTS to have better or worse play.
A key parameter to the MCTS algorithm is the number of {\it rollouts} used---the number of times the full cycle is repeated.
By increasing the number of rollouts allowed to an agent, we enable the agent to more fully explore the value of possible actions in the game and improve play~(see also Chapter 5, p.\ 60 in \cite{jaffe2013:thesis}).

We use MCTS rollouts as a proxy for player skill.
Modeling the effects of player skill offers several advantages to a design tool: 
many games are designed to reward more skilled players with greater rewards or higher win rates \cite{bjork2005:patterns};
designers are often concerned with differences in play style dependent on player skill \cite{elias2012:characteristics-games};
and games (including adversarial games) are often designed to enable a smooth progression of skill as players learn over time \cite{koster2013:theory-fun}.
%We use rollouts as a proxy for player skill, with more rollouts simulating a player that is better able to consider the outcomes of her actions in the game.
In adversarial games, varying the rollouts used by two MCTS agents can compare how gameplay looks when two agents vary in levels of skill, as well as compare the effects of relative differences in skill between two agents; e.g., comparing high-level play between two strong agents or comparing games between a weak and strong agent.
This is an improvement over human testing as it affords designers the ability to explore many different skill combinations, including some that may be difficult to examine using human playtesting alone. 
In our design metrics we compare differences in base player skill and differences between opposing players' skills.

\subsection{Game Domains}
Our case studies concern two games: the word game {\it Scrabble} and {\it Cardonomicon}, a competitive card game we developed that is inspired by {\it Hearthstone} and {\it Magic: The Gathering}.
Below we present these domains and discuss how each environment is represented to the MCTS agent.
%Move representations are important as they shape which actions an agent may take in the game, influencing the space of play explored by agents.
%Different move representations can yield different play patterns or result in wasted computation considering irrelevant moves (e.g., considering illegal moves that are rejected by the game).

%\begin{figure*}[tbh]
%\centering
%\includegraphics[height=1.5in]{./images/scrabblesque}
%\includegraphics[height=1.5in]{./images/cardonomicon}
%\caption{(a)~{\it Scrabble}, a word-creation tile game; (b)~{\it Cardonomicon}, a monster-based card game.}
%\label{fig:games}
%\vspace{-1.0\baselineskip}
%\end{figure*}

%%%%%%%%%%%%%%%%%%

\pseudosection{Scrabble.}
%note: discrete = tiles in game
%note: This feels a bit long and wordy, can probably cut down a bit
%
{\it Scrabble} is an adversarial game where players take turns placing tiles onto a game board to create words.
In {\it Scrabble}, players have a rack of seven tiles, each with a single letter, that is hidden from the opposing player.
In this implementation, we have simplified {\it Scrabble} so agents have perfect information about one another's states and perfect knowledge of all legal words.
On each player's turn, they select tiles from their rack and place them on the game board such that 1)~at least one of the tiles is placed adjacent to one of the other player's tiles and 2)~the tiles create dictionary words either left to right, top to bottom, or both.
The player that goes first, however, only needs to play a word that goes through the center space on the board.

%Moves in {\it Scrabble} are typically referred to in reference to tiles being placed on the board.
%Thinking about moves in this way can make it difficult for the MCTS agent to play the game since it would need knowledge about whether the tiles being placed actually form words. 
The MCTS agent represents moves on a turn as the word to form on that turn.  
Thus, the space of possible moves on a turn is all possible words that can be made on that turn.
After the current player has placed a word on the board, they receive points based on the letters used to form the word.
Each letter tile has a score associated with it; a word's score is the sum of the score values of the letters used to make that word.
The board is also populated with {\it bonus spaces} that increase the value of a word.
Bonus tiles available on a typical {\it Scrabble} board can double or triple the value of either a specific letter tile or of the word that the letter tile is part of.

Once a player receives points for a move, that player draws tiles at random until their rack is refilled with seven tiles and the turn ends.
Normally, the game ends when a player cannot draw new tiles and the winner is the player with the highest score at that point.
In our implementation, however, the the first player to meet or exceed 150 points wins.

%From a modeling standpoint, {\it Scrabble} is a relatively simple game.
%It is a turn-based game that takes place on a static game board and the actions available to players are discrete (forming words on the board).
{\it Scrabble} exposes several common design challenges when choosing content for player actions in a game.
Examples include determining a suitable distribution of tiles that players can choose from or determining the scores of individual tiles.
These factors are further complicated by the random tile drawing order.

%%%%%%%%%%%%%%%%%%

\pseudosection{Cardonomicon.}
{\it Cardonomicon} has the core elements of a class of game mechanic-heavy adversarial card games, exemplified by games like {\it Magic:\ The Gathering}.
From a design perspective, games like {\it Magic} are challenging to develop for several reasons. 
Each card must be balanced with respect to all other available cards: e.g., a single overly powerful card can make all other cards irrelevant.
Further, the random order of card draws and non-deterministic effects of actions introduce a large space of non-deterministic outcomes to sample over.
While {\it Magic} has hidden information, for simplicity {\it Cardonomicon} is perfect information.

In {\it Cardonomicon}, two players start with an identical deck of 20 cards representing creatures.
Gameplay consists of drawing cards, spending mana to place cards on the game board, and using cards to attack one another and the opposing player's hero.
Cards are parameterized by health, attack power, and mana cost.
Players start with a single hero card on the board with 20 health and 0 attack; a player loses when their hero's health is reduced to or below 0.
Each turn, players may play any combination of cards for which they can pay the mana costs.
A player's mana starts from 1 on the player's first turn and increases by 1 each turn up to a cap of 10.
Cards on the board may attack any other opposing card once per turn after the turn the card is played.
%Attacks reduce both the target and source minions' health by the attack of the opposing minion.
When a card attacks, the opposing card's health is reduced by the attacker's attack; attacking cards also receive damage from opposing cards.
When a card's health is at or below 0 it is removed from the board.
We designed a set of cards to allow the player to play one of multiple cards on each turn (with differing parameterizations), assuming they have drawn a playable card.

The MCTS agent in {\it Cardonomicon} represents moves as either playing a card or using a card to attack another card on the opponent's board.
One turn may involve multiple moves in a row.
The agent has one move for every card that can be played in the agent's hand and one move for every pair of their card attacking a target opponent card.
Only cards that may attack are represented and no attacks on the agent's own cards are permitted as this has no purpose in the {\it Cardonomicon} domain.
One additional move to end the turn is always available.
Thus, MCTS agents reason about whether to play a card, use a card to attack the opponent, or end their turn.

{\it Cardonomicon} exposes many common design challenges: ensuring all cards are worth using at some point during a game, avoiding degenerate strategies where players always follow the same rote actions, and understanding the space of decisions a player faces over the game.

%%%%%%%%%%%%%%%%%%%%%%%%%%%%%%%%%%%%%%%%%%%%%%%%%%%%%%%%%%%%%

\section{Skill-based Design Metrics}
In this work we focus on automating the analysis of game designs through design metrics for player actions and how they change based on player skill: e.g., more skilled players may score more points per turn in {\it Scrabble}.
We use a simple taxonomy to distinguish classes of design metrics: summaries, atoms, chains, and action spaces.
\begin{itemize}[noitemsep]
\item {\it Summaries} are high-level design metrics aggregating gameplay trace features of relevance to designers; e.g., the length of the game or probability of the first- or second-turn player winning.
\item {\it Atoms} are metrics tracking individual aspects of game state or actions; e.g., the frequency of playing a given letter tile or card.
\item {\it Chains} are gameplay patterns; we address two types: {\it combos}, regularities in how a player combines actions; and {\it counters}, regularities in how a player responds to opponent actions.\footnote{Our use of the terms `atom' and `chain' are distinct from those proposed by Dan Cook \cite{cook2007:chemistry-game-design}, but share the notion of distinguishing between single actions as atoms and patterned sequences of actions as chains.}
\item {\it Action spaces} represent the set of actions available to a player and how they vary over time; e.g., the number of unique cards a player can play on each turn.\footnote{Our use of action spaces is similar to Elias et al. `game arc' (Chapter 4, pp.\ 122--129 in \cite{elias2012:characteristics-games})}
\end{itemize}
Our taxonomy is not meant to be all-encompassing, but rather to offer organization to types of strategy design metrics and how they may be visualized and analyzed through simulated (or human) agents.
%While this taxonomy is not all-encompassing, it covers many relevant design metrics of interest, including the notions of ``game complexity tree'' \cite{elias2012:characteristics-games} and ``counter-play'' \cite{bjork2005:patterns, cadwell2013:counterplay} under chains 
Below we describe these metrics and ground them in our test game domains.

\pseudosection{Summaries.}
Summary analysis provides designers with high-level metrics to inform which other metrics to examine and how to frame the results.
In {\it Scrabble} and {\it Cardonomicon} summaries include the length of a game (in turns) and the probability of the first turn player winning.
%, or the probability of a player that plays a card wins the game.

\pseudosection{Atoms.}
Atoms characterize single actions in a game.
In {\it Scrabble}, atoms include the use of (or option to use) certain words to score points.
In {\it Cardonomicon}, atoms include playing cards and using cards to attack.
Analysis of atoms provides designers with an understanding of which actions and states in a game are being used (and which are not).
Getting a high-level sense of the gap between actual use of an action against when the action appears as an option enables designers to gauge whether specific actions are over-(or under-)used.
Comparing these results across agents with varying skill enables an understanding of which actions are relevant to high-(low-)level play.
Analysis can compare the rates of using different actions, having the option to use an action, or the difference between rates of using an action against the frequency of the action being available.
Visualizations of atoms include histograms to show frequency of use or availability of atoms.
%or line charts comparing atoms use or availability against the skill of an agent.

\pseudosection{Chains.}
Chains characterize recurrent gameplay patterns.
Chains subdivide among {\it combos}---action sequences used by a single player---and {\it counters}---action-counter pairs between actions taken by two players.
Combos appear in games with real-time action or where a player may take multiple actions within a single turn.
In {\it Cardonomicon}, combos include chains of playing cards or using cards to attack; {\it Scrabble} has no combos as each turn consists of a single move.
Counters appear in games with turn alternation or simultaneous moves where a player takes an action in response to an opponent action.
In {\it Scrabble}, counters involve playing a word in response to an opponent word choice; counters appear in {\it Cardonomicon} as attacking a card after an opponent plays it in the previous turn or playing a card in response to an opponent play.
Analyzing chains provides designers with insight into emergent strategy within a game, including chains of actions that may exercise a skill \cite{cook2007:chemistry-game-design} or ways players have discovered to thwart their opponents \cite{cadwell2013:counterplay}.
Understanding the strategies players use in a game can refine an understanding of what core gameplay loop players exercise (at varying levels of skill) to compare against a gameplay loop design goal.
Visualizations of chains include histograms of chains in a game and playtrace browsers showing sets of traces with chains highlighted as parts of those traces.

\pseudosection{Action Spaces.}
An action space characterizes atoms across time or across game states.
In {\it Scrabble}, an action space can visualize the number of distinct tiles played or the number of distinct words available to complete across turns in a game.
In {\it Cardonomicon}, an action space can visualize the number of distinct cards available to play or average number of cards able to attack across turns.
Action space analysis can provide information on the progression of a game and guide choices about pacing and growth of game complexity.
Designers can use action space visualizations to understand how the number of choices varies over the course of a game and how that range differs for players of differing skill.
Visualizations of actions spaces include line charts of action frequency over turns in a game, with multiple lines to indicate players of differing skill levels.

%%%%%%%%%%%%%%%%%%%%%%%%%%%%%%%%%%%%%%%%%%%%%%%%%%%%%%%%%%%%%

\section{Case Studies}
To demonstrate how player simulation and our metric taxonomy can aid in game design evaluation, we performed two case studies. 
The first case study of {\it Scrabble} explores how our metrics can evaluate a balanced game. 
The {\it Scrabble} case study verifies that our technique can identify balance in a design and differences in player skill.
The second case study of {\it Cardonomicon} shows how our metrics can assess a game with a flawed design. 
The {\it Cardonomicon} case study shows how simulated agents and design metrics identify game flaws to inform future design iterations. 

\subsection{Playtrace Collection}
Both studies used the same general methodology of sampling playtraces using MCTS agent pairs of varying computational bounds as a proxy for varying player skill. 
%For each game environment, we used simulated MCTS agents with varying rollouts to simulate users with varying levels of skill. 
%As our games are adversarial we parameterized rollouts as the {\it base rollouts} of the weaker of the two agents, and {\it rollout gap} as the number of additional rollouts used by the stronger agent compared to the weaker agent.
%For example, if the base rollouts was 100 and rollout gap was 100 the two agents used 100 and 200 rollouts.
%Parameterizing our experiments in this way allows a comparison of how gameplay occurs as both players increase in skill (base rollouts) and how gameplay varies when one agent has a greater skill advantage (rollout gap).
We varied agent reasoning to consider roughly one to two moves ahead in the game.
We used two moves ahead as an upper bound as research in reasoning on recursive structures suggests people are able to reason to roughly two levels of embedding; a result borne out in models of deductive reasoning on logic puzzles \cite{browne2013:deductive-search}.
Our MCTS selection policy (UCB1) forces trying all child moves of a given move once before repeating a move: thus all rollouts will first explore options for a single move before exploring two-move sequences.

To set computational bounds we approximated the average number of moves available to an agent and used this number to estimate the number of rollouts an agent would need to consider one or two moves ahead in the game.
To examine a range of agent capabilities we initially created three agent computational bounds: (1)~a {\it weak} agent with enough rollouts to explore the full set of moves on a given turn, but lacking resources to explore the two moves ahead, (2)~a {\it strong} agent with enough rollouts to fully explore moves on this turn and the next turn, and (3)~a {\it moderate} agent with rollouts halfway between these two.
Initial testing revealed little difference between the latter two agents; our results report agents that halve the number of rollouts of the two stronger agents as these more clearly illustrate the outcomes of variable player skill.
We believe the lack of differences derives from marginal returns for greater computational resources in our case study domains, likely due to their large branching factor.
%\todo{ref R2 question: do we need to say/do more?}

For each game domain we ran a pair of agents where each agent was set at one of these three levels.
For each agent pairing we simulated 100 games to get aggregate statistics on agent performance and visualized these results to examine relevant design metrics in both game domains.

In {\it Scrabble}, we approximated the number of rollouts for a single level deep by looking at the median number of possible words an agent could complete on a board: 50.
Thus, the weak agent used 50 moves.
Initially the strong agent was allowed 2500 rollouts ($50^2$ for two moves ahead) and the moderate agent 1250 rollouts.
After halving, this resulted in a moderate agent with 650 rollouts ($(1250 - 50)/2 + 50 = 650$) and a strong agent with 1250 rollouts.
%For the moderate and strong agents, we chose to vary how much exploration of the second turn would be done. 
%As such, we used 650 moves for the moderate agent and 1250 moves for the strong agent. 
%We did perform some experiments using 2500 moves (50 moves at one level and 50 moves for each of those moves), but found that many of the metrics produced by this agent were redundant with those produced by an agent using 1250 moves. 
%This is likely because we underestimated the actual space of possible moves and were not consistently sampling the full space of moves over two turns with the higher skilled agents. 

In {\it Cardonomicon}, we approximated the number card play options as choosing 2 cards to play each move out of a hand of 6 cards ($\binom{6}{2} = 15$ moves).
We modeled attack choices assuming the player (and opponent) has approximately 3 cards on the board and one hero card, yielding 3 source card choices for 4 targets ($3^4 = 81$ moves).
Together this yields a total of approximately 100 moves considered for the weak agent, 10000 for the strong, and 5000 for the modest.
After halving this resulted in 100, 2500, and 5000 rollouts for the weak, moderate, and strong agents, respectively.

%Candidate For Cut
Note that an alternative strategy to sampling up to two levels deep would be to have agents explicitly model a selection policy with pure exploration up to one or two levels.
In this case, search bounds would vary over the course of the game. 
We chose to use a fixed number of rollouts to capture the notion of agents of fixed `capability' in terms of resources to devote to reasoning.

%\begin{table}
%\centering
%\begin{tabular}{|c|c|c|c|}
%\hline 
% & {\bf Weak} & {\bf Moderate} & {\bf Strong} \\ 
%\hline 
%{\it Scrabble} & 50 & 650 & 1250 \\ 
%\hline 
%{\it Cardonomicon} & 100 & 2500 & 5000 \\ 
%\hline 
%\end{tabular}
%\label{tab:rollouts}
%\caption{MCTS agent rollouts in study domains.}
%\end{table}

\subsection{Scrabble Metrics}
Our first study shows how our metrics reveal balance and player skill differences in {\it Scrabble} despite the change to ending the game at 150 points.
The study shows these changes did not upset the game balance and demonstrates that {\it Scrabble} rewards higher skill play.
 
\begin{figure}[tb]
\centering
\includegraphics[height=2.25in]{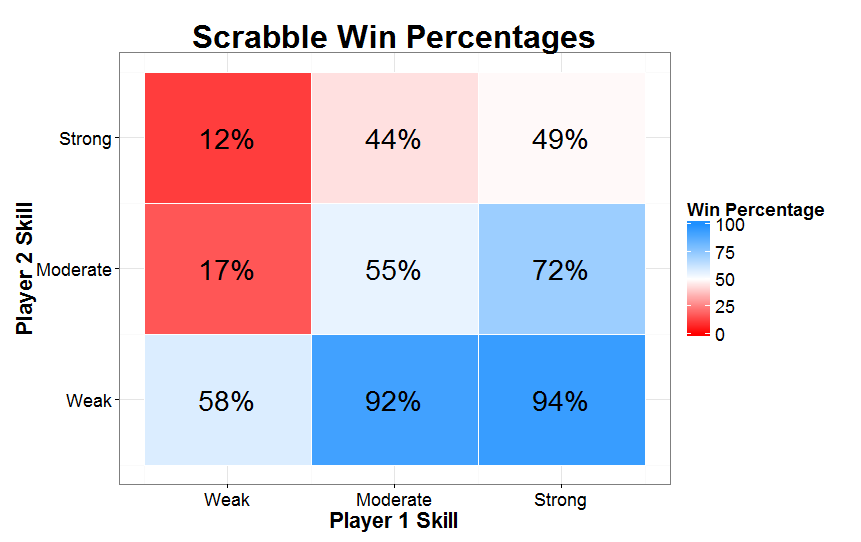}
\caption{Win percentage based on agent skill. Win percentages are calculated from the perspective of Player 1. Blue regions correspond to win percentage greater than 50\%. Red regions correspond to a win percentage less than 50\%.}
\label{fig:winPercentage}
\vspace{-1.0\baselineskip}
\end{figure}

\pseudosection{Summaries.}
The summary statistics that we examined in {\it Scrabble} are win percentage (Figure~\ref{fig:winPercentage}) and the length of a game based on turns.
Ideally, players with higher skill will consistently defeat lower-skilled opponents; however, it is unclear how skill will affect game length.

By comparing agents of varying rollouts we found the game is balanced: higher skilled agents consistently defeating lower skilled opponents (Figure~\ref{fig:winPercentage}).
This difference is very pronounced when the strong agent plays against the weak agent; however, it becomes less pronounced as agents increase in skill.
We also gathered metrics for first turn win-rate and found no first turn advantage. 
%This is shown by the strong agent having a slightly higher than 50\% win rate against the moderate agent. 

%\begin{figure}[tbh]
%\todo{AZ: replace me w/figure for card frequency; update reference to figure}
%\begin{figure}[tb]
%\centering
%\includegraphics[height=2.25in]{./images/scrabble_game_length_factored_skill.png}
%\caption{Game length based on agent skill. Game length is measured in turns. Darker regions correspond to longer games. Lighter regions correspond to shorter games.}
%\label{fig:turnLength}
%\vspace{-1.0\baselineskip}
%\end{figure}

We discovered that games played against skilled opponents are slightly shorter. 
When weak agents play against each other games last 26 turns on average; this decreases to 22 turns when strong agents play against each other.
This is likely because skilled opponents make moves worth more points. 
This belief is further supported by the results of our analysis of agent skill as it relates to the length of words played. 
Another possible explanation for this decrease in turns is that more skilled agents are able to better utilize the bonus tiles that exist on the board. 
Effectively using these tiles can drastically increase individual word scores. 
Analyzing the use of these tiles, however, showed there was little to no difference in usage patterns based on agent skill. 
Based on this finding, the main source of score difference between agents seems to stem from the length of words played.

\begin{figure}
\centering
\includegraphics[height=2.25in]{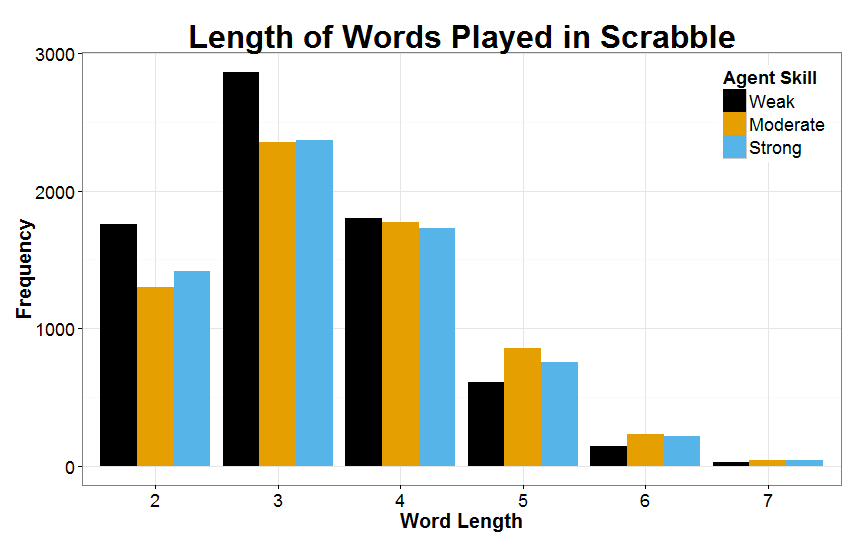}
\caption{Word length frequency in {\it Scrabble} by skill.}
\label{fig:wordDistribution}
%\vspace{-1.0\baselineskip}
\end{figure}

\begin{figure}
\centering
\includegraphics[height=2.25in]{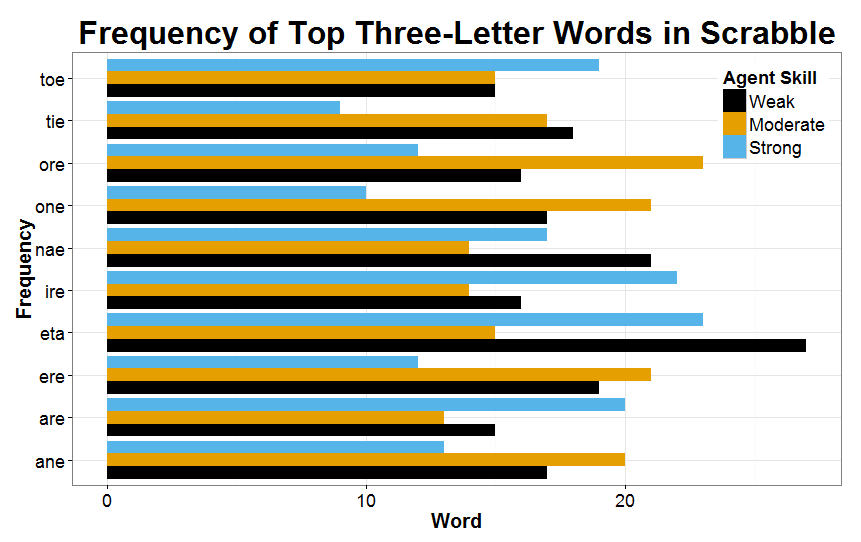}
\caption{Frequency of the top three-letter words in {\it Scrabble} by agent skill.}
\label{fig:top3LetterWords}
\vspace{-1.0\baselineskip}
\end{figure}

\pseudosection{Atoms.}
In {\it Scrabble}, the main atom metrics are from word usage rates as moves are words played.
Figure~\ref{fig:wordDistribution} shows the word usage distribution separated by word length and grouped by agent skill. 
Weak agents tend to favor playing shorter words, while stronger agents play a wider variety of word lengths. 
However, skill has little effect on the specific words played. 
Figure~\ref{fig:top3LetterWords} shows the most popular three-letter words in our simulations and how often each agent used each one. 
There is no consistent trend in the specific words an agent plays (while we show three-letter words, these findings were consistent across word lengths).

\pseudosection{Chains.}
% AZ: note - we defined `counter' so I changed that over for consistency
In {\it Scrabble}, {\it counters} are the words played by the opponent after a word has been played by the other player.
To determine what common counters in {\it Scrabble} were, we used frequent itemset mining on itemsets comprised of the words played on a given turn and the words played on the next turn.
Through this analysis, we discovered that most counters either add to the previously played word, or build a two or three-letter word off of the word that was previously played.
For example, one of the top counters to a player playing the word ``con'' on a turn was to add an ``i'' to the beginning of it to make the word ``icon.''
This is expected as building off previously played words will typically result in a higher point total since the player is playing a longer word than the opponent.

\pseudosection{Action Spaces.}
\begin{figure}[tb]
\centering
\includegraphics[height=2.15in]{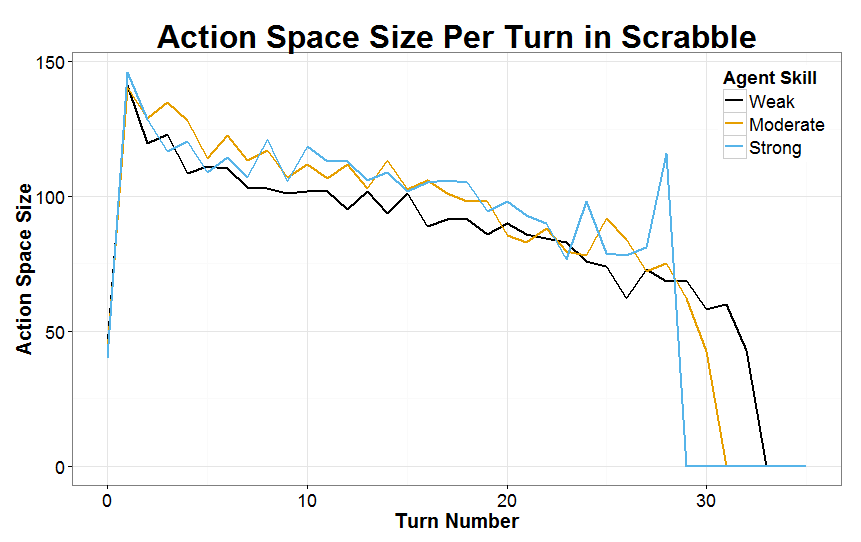}
\caption{Median number of words that could be played per turn based on skill.}
\label{fig:upperBound}
\vspace{-1.0\baselineskip}
\end{figure}
To describe the action space in {\it Scrabble}, we examined the space of possible words that can be played and were actually played. 
Figure~\ref{fig:upperBound} shows the median number of {\it possible} words that could have been played on a given turn based on skill. 
This conveys how the complexity of the action space changes over time. 
Figure~\ref{fig:upperBound} shows that the space of possible actions shrinks over the course of the game, likely because valid word placements become scarcer later in the game.
The figure also shows that stronger agents have more possible actions on a given turn than weaker agents.

\begin{figure}[tb]
\centering
\includegraphics[height=2.15in]{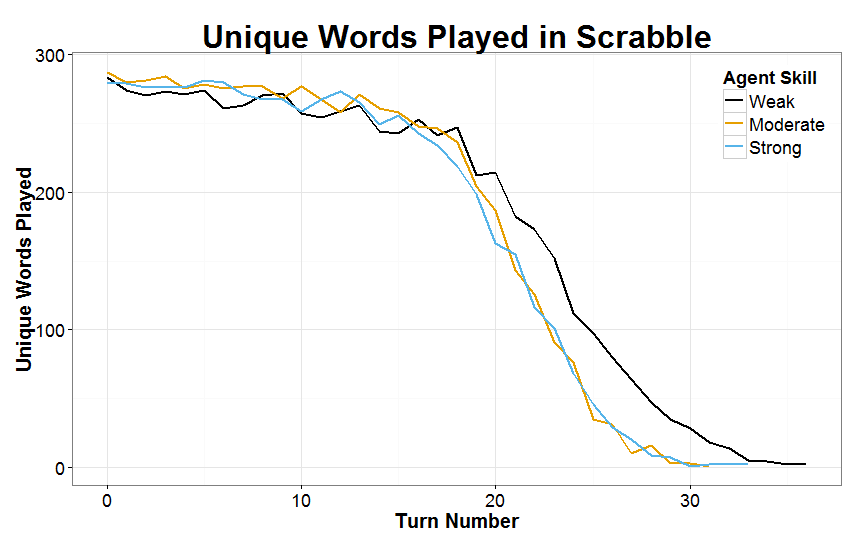}
\caption{Number of unique words played per turn by skill.}
\label{fig:uniqueActionsTaken}
\vspace{-1.0\baselineskip}
\end{figure}
Figure~\ref{fig:uniqueActionsTaken} shows how much of the action space was actually explored over the game. 
This figure shows that the space of words played shrinks faster for stronger agents than weaker agents, likely because stronger skilled agents successfully identify moves worth more points and avoid the rest of the action space.

%%%%%%%%%%%%%%%%%%%%%%%%%%%%%%%%%%%%%%%%%%%%%%%%%%%%%%%%%%%%%

\subsection{Cardonomicon Metrics}
Our second study examines {\it Cardonomicon}, showing how our metrics can identify design flaws. 
Recall that {\it Cardonomicon} is highly constrained in cards that are available to use and fixes the deck that players use. 
These major alterations to the typical structure of a card game negatively affected the balance of the game. 
In the following sections, we show this imbalance through our metrics. 

\begin{figure}[tb]
\centering
\includegraphics[height=2.25in]{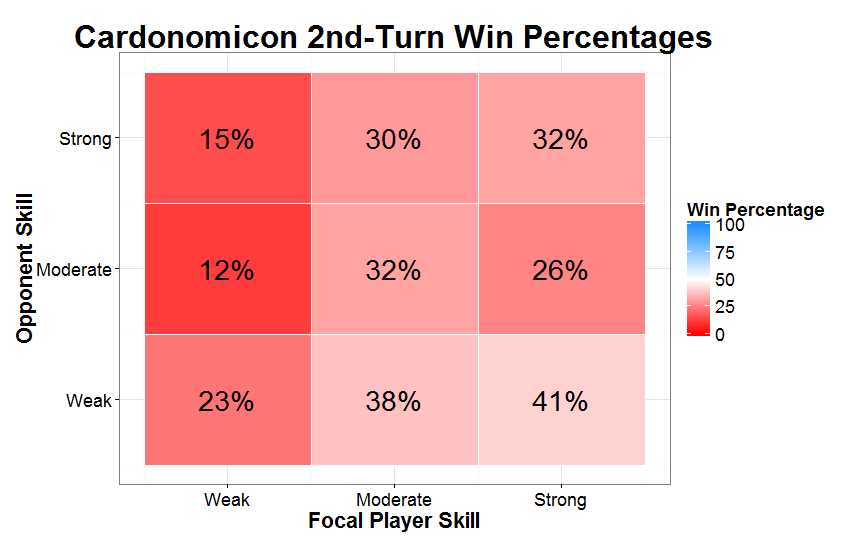}
\caption{Win rates for second turn player in {\it Cardonomicon}. The x-axis indicates agent strength for the second turn player; the y axis indicates the opposing agent's strength.}
\label{fig:cardt2winrate}
\vspace{-1.0\baselineskip}
\end{figure}
%\begin{table}[tb]
%\centering
%\begin{tabular}{c|ccc|}
<<<<<<< HEAD
=======
%\cline{2-4}
%  & \multicolumn{3}{c|}{opponent iterations} \\% \hline
>>>>>>> 2047ec2a6ea79c25ff8ee4bccdaf5d631ed606f7
%\cline{2-4}                                                                                      & \multicolumn{3}{c|}{opponent iterations} \\% \hline
%\multicolumn{1}{|c|}{\begin{tabular}[c]{@{}c@{}}focal player\\ iterations\end{tabular}} & 100 & 2500 & 5000 \\ \hline
%\multicolumn{1}{|c|}{100}                                                             & 23\% & 13\% & 15\% \\
%\multicolumn{1}{|c|}{2500}                                                            & 38\% & 32\% & 30\% \\
%\multicolumn{1}{|c|}{5000}                                                            & 41\% & 26\% & 32\% \\ \hline
%\end{tabular}
%\label{tab:cardt2winrate}
%\caption{Win rates for second turn player in {\it Cardonomicon}. Rows indicate agent strength for the second turn player; columns %are opposing agent strength.}
%\end{table}

\pseudosection{Summaries.}
A key design flaw in {\it Cardonomicon} is that the player going second is at a large disadvantage in terms of win rate.
Figure~\ref{fig:cardt2winrate} shows the win rates for the player who starts second.
Regardless of agent strength, the player going second has a win rate substantially less than 50\%.
That said, win rates do increase for the agent going second if they are more skilled than the agent going first. 
Thus, while agent skill influences player win rates in {\it Cardonomicon}, the game is flawed in giving a strong disadvantage to the player taking the second turn.
We speculate this is due to our partial adoption of mechanics from {\it Hearthstone}.
In {\it Cardonomicon}, cards are able to attack and receive damage in retaliation, but the second player has no advantage in being able to play more cards on their first turn.
As such, the second player will always be deploying cards after the first player, but lacks a mechanism to catch up to the player who acts first.

Stronger agents have (slightly) longer games when matched to evenly skilled opponents: median 16, 17, and 18 turns for the weak, moderate, and strong agents, respectively.
We attribute this trend to stronger agents being able to better counter one another while retaining enough cards to play until the end of the game.

\pseudosection{Atoms.}
{\it Cardonomicon} atoms consist of actions to play cards or use cards to attack.
When examining the frequency of playing different cards we found stronger agents generally play more cards, but show no large differences in their use of specific cards.
Stronger agents manage their mana to play more cards, but do not seem to favor specific cards to play.
This likely indicates the deck size in {\it Cardonomicon} is too small: agents will play all of their available cards faster than they draw new cards and thus have no opportunities to favor playing specific cards against others.%\footnote{Due to space concerns we have suppressed images that illustrate simple trends of the same form as shown with {\it Scrabble}.}

When examining the frequency of using cards to attack, stronger agents also tend to use cards to attack more overall.
Three cards showed disproportionately greater use by stronger agents compared to weaker agents: these three cards all had large amounts of health but low attack for their cost.
Strong agents thus use these cards to destroy multiple weaker cards by intelligently trading off card attacks and retaliations.
That is, stronger agents recognized the value in using a card with low attack (but high health) to remove several cards with lower attack and health over several turns.
This confirms {\it Cardonomicon} allows for a limited form of strategic variety and supports the notion that MCTS rollouts can help detect these potential strategic variants dependent on player skill.

\pseudosection{Chains.}
Chains in {\it Cardonomicon} are primarily combos: sequences of actions taken by a single player in a turn of the game.
As expected from our results from the atom analysis, we found no significant combos in terms of playing or attacking cards. 
We attribute this to the lack of any strong synergy among cards in {\it Cardonomicon}: no pairs were particularly outstanding as no pairs had effects that would be advantageous to use together.
This highlights another way to detect design flaws through these metrics: the absence of chains indicates no strong synergies exist in the design for players to use in combos.

\begin{figure}[tb]
\centering
\includegraphics[height=1.75in]{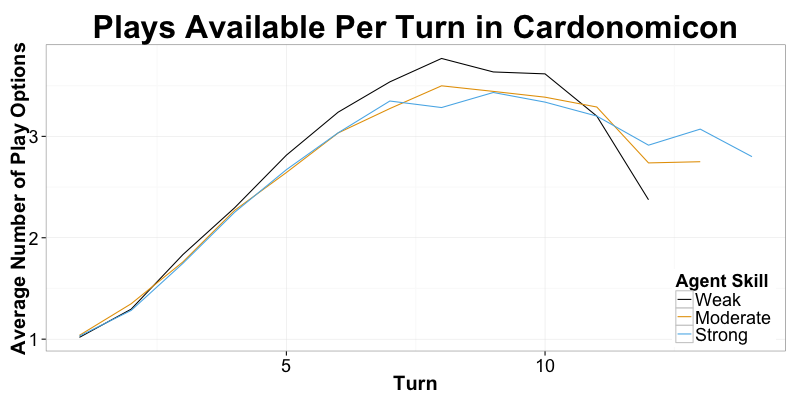}
\caption{Average number of possible card plays per turn based on skill.}
\label{fig:cardonomicon-playspace}
\vspace{-1.0\baselineskip}
\end{figure}

\begin{figure}[tb]
\centering
\includegraphics[height=1.75in]{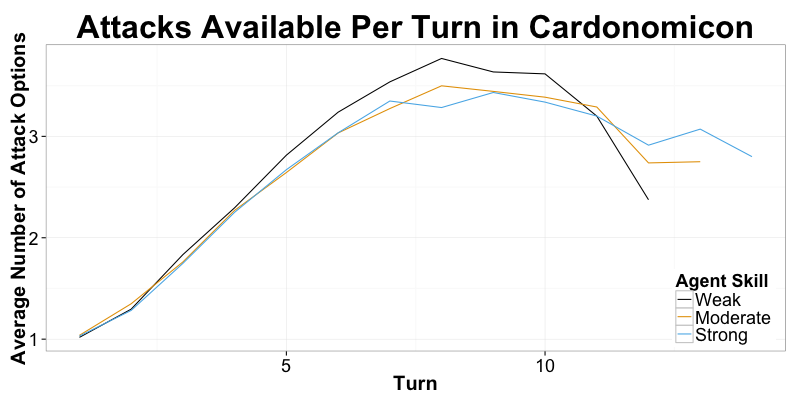}
\caption{Average number of possible attacks per turn based on skill.}
\label{fig:cardonomicon-attackspace}
\vspace{-1.0\baselineskip}
\end{figure}

\pseudosection{Action Spaces.}
As with {\it Scrabble}, stronger {\it Cardonomicon} agents have a larger space of cards they may play (Figure \ref{fig:cardonomicon-playspace}) and use to attack (Figure \ref{fig:cardonomicon-attackspace}).
Specifically, we observed stronger agents have more options to play cards late in the game, while having fewer mid-game attack options with more late-game attack options.
These results align with intuition: in the early game both weak and strong players have a similar range of options constrained primarily by the amount of mana players have.
By mid-game stronger players will have fewer attack options as they retain cards they may play for the late game.
Playing these cards in the late game leads to more options to attack.
Aligning with these analyses of the number of {\it possible} plays, we observed that more skilled players both play and attack with a larger number of cards on average. 
They also are capable of a larger number of possible actions in these game phases.
Thus, skilled players also actually use this larger set of options.
Overall, these results demonstrate that more skilled players in {\it Cardonomicon} will open more plays in the mid-game by intelligently retaining cards before using these cards in the late game; in sum, these players are more efficient in their use of mana.
%options:
%	CN_actionspace_play_option.pdf
%		- higher skilled have more late game play options
%	CN_actionspace_abl_option.pdf
%		- higher skilled still have attack options late game by saving from mid-game
%
%average plays/attacks used:
%	CN_actionspace_card_play.pdf
%		- higher skill shows increase in cards played by late game (same early)
%	CN_actionspace_card_abl.pdf
%		- higher skill shows (noisy) shift from mid -> late game attack options
%
%actions taken per turn:
%	CN_space_play.png
%		- higher skilled save more plays for late game
%	CN_space_abl.png
%		- higher skilled have more attacks available late game
%
%overall: more skilled players play more unique cards, enabling more unique abilities
%	- early play is similar
%	- midgame opens more plays
%	- late game has more plays, more attacks
%	-> indicates better mana utilization

%CN_actionspace_abl_option
%CN_actionspace_play_option

%%%%%%%%%%%%%%%%%%%%%%%%%%%%%%%%%%%%%%%%%%%%%%%%%%%%%%%%%%%%%

\section{Limitations}
Our technique has two key limitations: (1)~MCTS rollouts only represent one dimension of player skill and (2)~the algorithms used only apply to fully observable domains.
Varying MCTS rollouts alters player skill in terms of the extent to which an agent considers different courses of action.
Other aspects of skill in games---including visual perception, motor control, memorization and recall, etc.---are not modeled by rollouts.
Thus, in {\it Scrabble} our approach assumes all players have perfect knowledge of game systems in terms of valid dictionary words and always apply this knowledge.
Generalizing our technique to real-time domains will require modeling other facets of player skill including visual perception and motor control.
Real-time MCTS variants may apply to these cases, though determining the appropriate skill proxies is an open problem \cite{browne2012:mcts-survey}.

Both of our case study domains have been simplified to be fully observable.
Algorithms for MCTS that address domains with partial information can potentially be applied as substitutes for the simple implementation we use \cite{browne2012:mcts-survey}.

%%%%%%%%%%%%%%%%%%%%%%%%%%%%%%%%%%%%%%%%%%%%%%%%%%%%%%%%%%%%%

\section{Conclusion}
In this paper, we have shown how anytime planning agents can simulate players of various skill levels in turn-based, adversarial games.
We provide several types of metrics to analyze (human or simulated) player strategies from playtraces---at the levels of summaries, atomic actions, action chains, and action spaces---and show how these metrics can identify balance concerns and differences in player strategic play.
%These metrics capture common aspects of strategic analysis in game design and render them into computationally evaluated metrics and visualizations.
We show the value of these metrics by using them to assess two different games: {\it Scrabble} and the card game {\it Cardonomicon}.
In {\it Scrabble}, these metrics show the game remained balanced despite our minor alterations to it.
In addition to identifying balance, our case study on {\it Cardonomicon} shows these metrics can help identify design flaws.
The ability to identify imbalances and differences in player strategies provides designers with new lenses on game design that ground concepts often discussed in design analyses \cite{elias2012:characteristics-games, salen2003:rulesplay}.
Through simulated design evaluation systems we hope to augment the practice of game design, enable the scientific testing of game design hypotheses in terms of how designs influence player strategy, and improve the state of automated game generation systems through more sophisticated design evaluation.

%How on earth to bring this home...
%This work represents a step towards moving beyond using computers as tools to create

\section{Acknowledgments}
%IML money?
%ACT money?
We would also like to thank our reviewers for highlighting additional related work and noting abstraction as a technique to augment sampling approaches.

\bibliographystyle{abbrv}
\bibliography{lib}

\end{document}